\documentclass{article}

\usepackage[final,nonatbib]{neurips_distshift_2022}
\usepackage{authblk}
\usepackage[utf8]{inputenc} 
\usepackage[T1]{fontenc}    
\usepackage{hyperref}       
\usepackage{url}            
\usepackage{booktabs}       
\usepackage{amsfonts}       
\usepackage{nicefrac}       
\usepackage{microtype}      
\usepackage{xcolor} 
\usepackage{amsmath}
\usepackage{graphicx}
\usepackage{listings}
\usepackage{floatrow}
\newfloatcommand{figurebox}{figure}[\nocapbeside][\dimexpr(\textwidth-\columnsep)/2\relax]
\newfloatcommand{tablebox}{table}[\nocapbeside][\dimexpr(\textwidth-\columnsep)/2\relax]
\title{A Simple Baseline that Questions the Use of Pretrained-Models in Continual Learning }

\author[1,3 \thanks{Workdone during internship at KAUST}]{Paul Janson}
\author[1]{Wenxuan Zhang}
\author[2]{Rahaf Aljundi}
\author[1]{Mohamed Elhoseiny}
\affil[1]{King Abdullah University of Science and Technology, Saudi Arabia}
\affil[2]{Toyota Motor Europe, Belgium}
\affil[3]{University of Moratuwa,
Sri Lanka}

\begin{document}

\maketitle
\vspace{-2mm}
\begin{abstract}
With the success of pre-training techniques in representation learning, a number of continual learning methods based on pre-trained models have been proposed.  Some of these methods design continual learning mechanisms on the pre-trained representations and only allow minimum updates or even no updates of the backbone models during the training of continual learning. In this paper, we question whether the complexity of these models is needed to achieve good performance by comparing them to a very simple baseline that we design. We argue that the pre-trained feature extractor itself can be strong enough to achieve a competitive or even better continual learning performance on Split-CIFAR100 and CoRe 50 benchmarks.  
To validate this, we conduct the baseline that 1) uses a frozen pre-trained model to extract image features for every class and computes their corresponding mean features during the training time, and 2)  makes predictions based on  the nearest neighbor distance between test  samples and mean features of the classes; i.e., Nearest Mean Classifier (NMC). This baseline is single-headed, exemplar-free, and can be task-free by updating the mean features continually. This baseline achieved $83.70\%$ on 10-Split-CIFAR-100, surpassing most state-of-the-art continual learning methods, with all initialized by the same pre-trained transformer model. We hope our baseline may encourage future progress in designing learning systems that can continually add quality to the learning representations even if they start from pre-trained weights. Code is available  \url{https://github.com/Pauljanson002/pre-trained-cl.git}
\vspace{-5mm}
\end{abstract}

\vspace{-2mm}
\section{Introduction}
Conventional machine learning models struggle to perform well when the i.i.d. assumption is violated in the real world, where data arrives sequentially from tasks with shifting distributions over time. These models suffer from catastrophic forgetting of earlier tasks\cite{mccloskeyCatastrophicInterferenceConnectionist1989,kirkpatrick_overcoming_2017}. 
Continual learning, in which the agent is expected to learn new tasks without forgetting the old ones, has been studied extensively as a potential solution to this issue.  Early approaches start training with a model from scratch,  continually adapt the model for future tasks, and prevent forgetting by replaying the data, designing the penalization of the updates of the model parameters, and/or dynamically increasing model parameters to incorporate new knowledge.

The use of pre-training \cite{steinerhow} on large-scale datasets, such as ImageNet-1k\cite{russakovsky_imagenet_2015} and ImageNet-21k \cite{ridnik_imagenet-21k_2021}  has led to significant advances.  With sufficient and diverse data, the output features of a pre-trained model generalize well to a vast array of tasks,  greatly increasing the performance of challenging learning scenarios such as continual learning. As a result, some existing works directly deploy the pre-trained models as feature extractors, and apply continual learning techniques  on the feature level. These methods imply that pre-trained models provide general and coarse features that need to be task-specifically fine-tuned. However, \cite{radford2021learning} stated large-scale pre-trained models have excellent classification performance to the data outside of the pre-trained distribution even without extra training. We are questioning whether these pre-trained features are competitive for downstream tasks and  whether sophisticated continual learning techniques are indeed needed to achieve a good performance on the studied continual learning benchmarks.


To answer these questions, we implement a simple baseline in continual learning. We use a frozen pre-trained model to extract features for the training set and compute mean features to represent each class. During the testing time, we make predictions based on the distance between the test samples and class mean features. Our results are surprisingly competitive with SOTA methods using the same pre-trained backbone network. We achieve $83.70$\% Average  Accuracy at the end of the learning sequence of  Split CIFAR100 compared to $83.83\%$ of L2P\cite{wang_learning_2022}, and $83.23\%$  on the evaluation task of CoRe50 compared to $78.33\%$ of L2P. This implies that the pre-trained models provide quality and robust learning representations  under distribution shifts. We also present the experimental results on two more diverse benchmarks, 5-dataset and Split-ImageNet-R. 
The results on these datasets also show competitive performance outperforming most methods. 


Our proposed baseline is single-headed, task-free, and exemplar-free as an additional bonus. We argue that this is a simple yet powerful baseline that every continual learning method should compare against.  The intentions of these properties are to make less use of task labels (which is unrealistic in practice) and replay data (which raises privacy concerns) in the techniques. We observe that some prior works  outperform this baseline, but at the sacrifice of one or more of these characteristics. There are few comparable approaches with ours when all the conditions align. We hope that this work sheds some light on examining the practicality of pre-training in continual learning and whether the new methods are improving the learned representation quality continually.

\vspace{-3mm}
\section{Related Work}
\vspace{-2mm}
\textbf{Continual learning with pre-trained models }
Continual learning methods generally trained feature extractors from scratch and constrained the drift in feature representation \cite{kirkpatrick_overcoming_2017,mccloskey_catastrophic_1989}. Recently, the usage of pre-trained models has attracted more attention in continual learning. \cite{ferrari_end--end_2018}  viewed the first task as a pre-training stage and froze the feature representation after the first task. \cite{gallardo_self-supervised_nodate-1} found that a larger size of data in the first task and self-supervised pre-training helped to decrease catastrophic forgetting. \cite{lesort_scaling_2022} empirically analyzed the effect of training the last layer with a fixed feature extractor. \cite{ostapenko_continual_2022} studied foundation models and replay of frozen latent features to overcome catastrophic forgetting. Recently \cite{wang_learning_2022} proposed to use prompt based fine-tuning with a pre-trained transformer \cite{dosovitskiy2020image,vaswani2017attention}  and compared it with other methods in the same initialization. We adopt this strategy and point out that classifier learning actually reduced the power of representations learned by the pre-trained model.

\textbf{Task-aware/free continual learning}
Task incremental method requires the task identifiers of the samples during the inference, which reduces the practical application of these methods,  whereas class-incremental methods do not require those.  However,  task-aware/free methods suffer from class recency bias. To overcome such recency bias, BiC\cite{wu_large_2019} proposed to add a final layer that reduces the final bias. LUCIR \cite{hou_learning_2019} proposed to use a balanced fine-tuning to train the classifier after freezing feature representation. Our method also focuses on this problem and uses the simple nearest mean classifier on top of pre-trained transformer.

\textbf{Exemplar-free continual learning}
Earlier methods were proposed to store raw data, learned features, or generated features from previous tasks. Since the replay mechanism such as ER \cite{chaudhry_tiny_2019} is orthogonal to architecture-based and regularization-based methods, they are widely adopted in other methods to improve performance.  Recently there has been an increased interest in exemplar-free continual learning to account for privacy concerns in storing raw samples  and  storage concerns.  iCaRL \cite{rebuffi_icarl_2017} introduced the method of using exemplars for continual learning and selected exemplars close to the class means. Our baseline follows a similar  strategy but only stores the class-mean in feature space which save the  storage  by reducing the number of latent variables to keep per class.

\vspace{-2mm}
\section{Methodology}

 \label{sec:problem_setup}
\textbf{Problem Setup} We adopt the standard continual learning scenario where a model  learns from a non-i.i.d. data stream, represented as $\mathcal{D}_1,...,\mathcal{D}_{T}$, where $\mathcal{D}_{t} = \{(x_{i}^{t},y_{i}^{t}) \}_{i=1}^{N_t}$ is the task-specific subset,  $x_{i}^{t} \in \mathbb{R}^{w \times h \times c}$ is an image input and $y_{i}^{t} \in \mathbb{Z}$ is its corresponding label. 
The goal of continual learning is to learn a function $f_{\theta}$ which maps the input $x$ to the label $y$ from an arbitrary task seen so far. 
We focus on two scenarios. 
In the class-incremental setting, each subset $\mathcal{D}_t$ contains a disjoint class label set. In the domain-incremental setting, the subsets $\mathcal{D}_1, \dots, \mathcal{D}_T$ share the class labels, but the  input distributions varies over time. 

\textbf{Nearest Mean Classifier (NMC)}
We decouple the goal of continual learning $f_{\theta}$ into two steps. The first step is to learn the representation $h$ and the next is to learn the classifier $g$. We directly adopt a pre-trained vision transformer  as our feature representation without training. For the classifier, inspired by \cite{rebuffi_icarl_2017,mensink_distance-based_2013-1} we use the nearest mean classification strategy. 
During the training stage of task $t$, we calculate the mean features of a class in $\mathcal{D}_t$  
\begin{equation}
\label{eq:mean_calculation}
    \mu_{k} = \frac{1}{|C_{k}|} \sum_{x \in C_{k}} h(x) \enspace,
\end{equation}
where $C_{k}$ denotes the set of training samples belonging to class $k$.
Only class mean features are saved in the memory and will be used during evaluation. At the test time of task $t$, the feature of a test sample  is extracted by the pre-trained model, and the predicted class label is taken as the class whose mean features is the closest (over all the seen classes so far) to the feature of a test sample. 
\begin{equation}
    \hat{y} = \underset{k}{\text{argmin}} || h(x) - \mu_{k} || 
\end{equation}
\vspace{-5mm}




\vspace{-2mm}
\vspace{-2mm}
\section{Experiments}
\vspace{-2mm}

We follow the experimental setup used in \cite{wang_learning_2022} to evaluate our method for a fair comparison. We test our method in class incremental learning, where new sets of classes are introduced to the model, and in domain incremental learning, where classes remain the same and the domain changes; see Sec \ref{sec:problem_setup}. 

\textbf{Datasets: }
We evaluate our baseline on four common continual learning benchmarks, Split- CIFAR100 \cite{krizhevsky2009learning} , 5-datasets \cite{ebrahimi_adversarial_2020} and Split-ImageNet-R \cite{hendrycks2021many} in the class incremental learning setting. As proposed by \cite{wang_learning_2022} and \cite{wang_dualprompt_2022}. Split-CIFAR-100 contains 10 tasks with 10 classes for each task.
The 5-dataset benchmark concatenates 5 datasets, MNIST, SVHN, notMNIST , FashionMNIST and CIFAR10,  with each dataset forming one task. Split ImageNet-R is a newly proposed dataset for continual learning by \cite{wang_dualprompt_2022}. It consists of 200 classes which are randomly divided into 10 tasks. It contains the same  object types however presented in different styles such as cartoon , graffiti and origami. These variations make  the continual learning more challenging.
For the domain incremental learning setting, we use CoRe50 proposed by \cite{lomonaco2017core50}. It contains 50 objects collected in 11 distinct domains(tasks). 8 domains were  faced and learned incrementally while the test is performed on the remaining three domains. Since a single test task is used, we do not report forgetting and joint training results in that scenario.


\textbf{Evaluation Methods: } For our approach (\texttt{Ours}), we employ the widely used ViT-B/16\cite{dosovitskiy2020image} model pre-trained on ImageNet-21k \cite{steinerhow} provided by \verb|timm| library \cite{rw2019timm}. We mainly compare our baseline with the  recent L2P \cite{wang_learning_2022} which adopts the pretrained model as ours  and learns a prompt pool with a prompt selection mechanism to modify the pretrained representations.  We also consider  popular continual learning methods including regularization-based methods (LwF\cite{li_learning_2018} , EWC \cite{kirkpatrick_overcoming_2017} ) and rehearsal-based methods (ER\cite{chaudhry_tiny_2019}, GDumb\cite{vedaldi_gdumb_2020}, BiC \cite{wu_large_2019}, DER++ \cite{buzzega_dark_2020} and Co$^{2}$L \cite{cha2021co2l}).  We  present joint training results where the training data is, i.i.d. distributed among the whole benchmark with no task split. FT-frozen adds a fully-connected layer on top of the frozen feature extractor as the classification head, and FT allows end-to-end training on the feature extractor. Note that FT-frozen is different from our baseline, as we use NMC classifier and build it incrementally.

\begin{table}[t!]
\vspace{-4mm}
    \begin{floatrow}[2]
    \tablebox[\FBwidth]{\caption{Continual learning performance expressed in Average Accuracy and Forgetting at the end of the learning sequence  of CIFAR-100\cite{krizhevsky2009learning}. All methods are initialized with pretrained weights for a fair comparison. Our baseline shows competitive performance on this benchmark. }
\label{tab:cifar100}}{
\resizebox{!}{1in}{%
\begin{tabular}{@{}cccc@{}}
\toprule
Method        & Buffer size & Average Acc    & Forgetting \\ \midrule
FT - frozen   & 0           & 17.72          & 59.09      \\
FT            & 0           & 33.61          & 86.87      \\
EWC\cite{kirkpatrick_overcoming_2017}           & 0           & 47.01          & 33.27      \\
LwF \cite{li_learning_2018}          & 0           & 60.69          & 27.77      \\
L2P  \cite{wang_learning_2022}         & 0           & \textbf{83.83}          & 7.63       \\
\textbf{Ours} & 0           & 83.70 & -          \\ \midrule
ER  \cite{chaudhry_tiny_2019}          & 50/class    & 82.53          & 16.46      \\
GDumb \cite{vedaldi_gdumb_2020}        & 50/class    & 81.67          & -          \\
BiC \cite{wu_large_2019}           & 50/class    & 81.42          & 17.31      \\
DER++ \cite{buzzega_dark_2020}        & 50/class    & 83.94          & 14.55      \\
Co2L \cite{cha2021co2l}         & 50/class    & 82.49          & 17.48      \\
{L2P} \cite{wang_learning_2022}  & 50/class    & {86.31} & 5.83       \\ \midrule
Joint         & -           & 90.85          & -          \\ \bottomrule
\end{tabular}%
}
}
\tablebox[\FBwidth]{\caption{Continual learning performance expressed in Average Accuracy and Forgetting at the end of the learning sequence of the 5-dataset benchmark \cite{ebrahimi_adversarial_2020}. All methods are initialized with pretrained weights of the transformer for a fair comparison. Our baseline performs competitively with the exemplar-free methods in this benchmark}\label{tab:5datasets}
}{
\resizebox{!}{1in}{
\begin{tabular}{@{}cccc@{}}
\toprule
Method        & Buffer size          & Average Acc    & Forgetting           \\ \midrule
FT - frozen   & 0                    & 39.49          & 42.62                \\
FT            & 0                    & 20.12          & 94.63                \\
EWC\cite{kirkpatrick_overcoming_2017}           & 0                    & 50.93          & 34.94                \\
LwF \cite{li_learning_2018}           & 0                    & 47.91          & 38.01                \\
L2P \cite{wang_learning_2022}           & 0                    & \textbf{81.14}          & 4.64                 \\
\textbf{Ours} & 0                    & 79.84 & -                    \\ \midrule
ER \cite{chaudhry_tiny_2019}           & 50/class             & 84.26          & 12.85                \\
GDumb \cite{vedaldi_gdumb_2020}        & 50/class             & 70.76          & -                    \\
BiC \cite{wu_large_2019}          & 50/class             & 85.53          & 10.27                \\
DER++ \cite{buzzega_dark_2020}        & 50/class             & 84.88          & 10.46                \\
Co2L \cite{cha2021co2l}         & 50/class             & 86.05          & 12.28                \\
{L2P}\cite{wang_learning_2022}  & 50/class             & \textbf{88.95} & 4.92                 \\ \midrule
Joint         & \multicolumn{1}{l}{} & 93.93          & \multicolumn{1}{l}{}\\\bottomrule
\end{tabular}
}
}
    \end{floatrow}
\end{table}

\begin{table}[h!]
\vspace{-2mm}
\label{tab:core50}
\centering
\begin{floatrow}[2]
\tablebox[0.5\textwidth]{\caption{Continual learning performance in Average Accuracy at the evaluation task of CoRe50 \cite{lomonaco2017core50}. All methods are initialized with pretrained weight for fair comparison}}{
\resizebox{!}{0.9in}{%
\begin{tabular}{ccc}
\toprule
Method & Buffer size & Test Acc \\ \midrule
EWC\cite{kirkpatrick_overcoming_2017} & 0 & 74.82 \\
LwF\cite{li_learning_2018} & 0 & 75.45 \\
L2P\cite{wang_learning_2022} & 0 & 78.33 \\
\textbf{Ours} & 0 & \textbf{83.23} \\ \midrule
ER\cite{chaudhry_tiny_2019} & 50/class & 80.1 \\
GDumb\cite{vedaldi_gdumb_2020} & 50/class & 74.92 \\
BiC\cite{wu_large_2019} & 50/class & 79.28 \\
DER++\cite{buzzega_dark_2020} & 50/class & 79.7 \\
Co2L\cite{cha2021co2l} & 50/class & 79.75 \\
L2P\cite{wang_learning_2022} & 50/class & 81.07 \\ \bottomrule
\end{tabular}%
}
}
\tablebox[0.5\textwidth]{\label{tab:imagenet}\caption{Continual learning performance in Average Accuracy and Forgetting at the end of the learning sequence of the Split-ImageNet-R\cite{hendrycks2021many} benchmark. All methods are initialized with pretrained weights for fair comparison.}
}{
\resizebox{!}{0.9in}{
\begin{tabular}{@{}cccc@{}}
\toprule
Method        & Buffer size & Average Acc.   & Forgetting \\ \midrule
FT - frozen   & 0           & 39.49          & 42.62      \\
FT            & 0           & 28.87          & 63.80      \\
EWC \cite{kirkpatrick_overcoming_2017}          & 0           & 35.00          & 56.16      \\
LwF  \cite{li_learning_2018}         & 0           & 38.54          & 52.37      \\
L2P \cite{wang_learning_2022}          & 0           & 61.57          & 9.73       \\ 
\textbf{Ours} & 0           & 55.56         & -          \\\midrule
ER   \cite{chaudhry_tiny_2019}         & 5000        & 65.18          & 23.31      \\
GDumb  \cite{vedaldi_gdumb_2020}       & 5000        & 65.90          & -          \\
BiC  \cite{wu_large_2019}         & 5000        & 64.63          & 22.25      \\
DER++ \cite{buzzega_dark_2020}        & 5000        & \textbf{66.73} & 20.67      \\
Co2L \cite{cha2021co2l}         & 5000        & 65.90          & 23.36      \\ \midrule
Joint         &    -         & 79.13          &  -      \\\bottomrule
\end{tabular}}
}
\end{floatrow}

\vspace{-5mm}

\end{table}
\textbf{Results: }
Table~\ref{tab:cifar100}, \ref{tab:imagenet}, and \ref{tab:5datasets} report the performance of continual learning in incremental setting in Average Accuracy at the end of the learning sequence of Split-CIFAR100, Split-ImageNet-R, and the 5-dataset  respectively. Table~\ref{tab:core50} shows the results of the continual learning in domain incremental setting on CoRe50\cite{lomonaco2017core50}. The results are grouped based on the use of replay sample. 
Our simple baseline achieves the competitive performance on Split CIFAR-100 and CoRe50 , Split-ImageNet-R and 5-dataset benchmarks. Our results on Split-CIFAR-100 are even better than  methods that use replay samples. Concretely, our baseline achieves 83.70\% with  zero buffer size.
This suggests that pre-trained transformer offers a strong representation that achieves competitive performance. We think that the main reason for a possible inferior performance could be the ineffective design of the continual learning mechanisms compared to the robust features provided by the pretrained model. Such a model,  pretrained on a large and diverse dataset, may have already captured most of the distribution properties in the evaluation benchmarks. Then a desired continual learning mechanism needs to further encourage the model to produce the task-invariant features specific to the deployed benchmark. 
Our nearest neighbor baseline  is also competent among exemplar-free methods on Split-ImageNet-R and  5-datasets  benchmarks, which is more diverse than CIFAR-100 and CoRe50. However, methods with higher buffer sizes and finely designed continual learning mechanisms do improve the representations extracted from pretrained models. 
\vspace{-2mm}


\vspace{-2mm}
\section{Conclusion}

In this work, we explore the representational capacity of large-scale pre-trained models in continual learning settings. 
We provide simple nearest neighbor  baseline experiments on four benchmarks, showing  competitive performance to more sophisticated state-of-the-art continual learning methods which also leverage the same pretrained models. 
We agree that using pretrained weights can be a reasonable practice even in continual learning.  However, to show real progress in continual learning systems, we need to focus more on building  methods that can continually add quality to the learning representations. 
A desired continual learning algorithm shall go significantly beyond the knowledge embedded in the pretrained model.
Another important aspect is the considered benchmarks for evaluating continual learning methods. Such benchmarks need to be sufficiently challenging and   different from the data distributions employed for pretrained models. 

\medskip
{
\bibliography{main}

}
\end{document}